\documentclass[11pt,a4paper]{article}
\usepackage[hyperref]{eacl2021}
\usepackage{times}
\usepackage{latexsym}

\usepackage{booktabs}

\usepackage{microtype}
\usepackage{multirow}
\usepackage{graphicx}
\usepackage{subcaption}
\usepackage{esvect}
\usepackage{comment}

\newcommand{\bgcomment}[1]{\textcolor{cyan}{Bhanu: #1}}
\newcommand{\B}[1]{\textbf{#1}}

\usepackage{microtype}

\aclfinalcopy 

\title{\textsc{EmpathBERT}: A BERT-based Framework \\ for Demographic-aware Empathy Prediction}

\author{Bhanu Prakash Reddy Guda \\
  Adobe Research \\
  \texttt{guda@adobe.com} \\\And
  Aparna Garimella \\
  Adobe Research \\
  \texttt{garimell@adobe.com} \\\And
  Niyati Chhaya \\
  Adobe Research \\
  \texttt{nchhaya@adobe.com} \\
  }
  
\date{}

\begin{document}
\maketitle
\begin{abstract}
Affect preferences vary with user demographics, and tapping into demographic information provides important cues about the users' language preferences.
In this paper, we utilize the user demographics, and propose {\it \textsc{EmpathBERT}}, a demographic-aware framework for empathy prediction based on BERT. 
Through several comparative experiments, we show that {\sc EmpathBERT} surpasses traditional machine learning and deep learning models, and illustrate the importance of user demographics to predict empathy and distress in user responses to stimulative news articles.
We also highlight the importance of affect information in the responses by developing affect-aware models to predict user demographic attributes.
\end{abstract}

\section{Introduction}

Modeling complex human reactions and affect from text has been a challenging research area with innovations focusing on sentiment and emotion understanding \cite{Picard1997,li-liu-2015-improving, Rosenthal2017, Socher2011, Socher2013}. The study of non-trivial human reactions has been limited. These methods, often rooted in psychological theories, have turned out to be more complex in terms of annotation and modeling \cite{strapparava-mihalcea-2007-semeval}.  A critical affective phenomena, \textit{empathy}, has received surprisingly less attention.

\begin{figure}
    \centering
    \includegraphics[width=1.0\columnwidth]{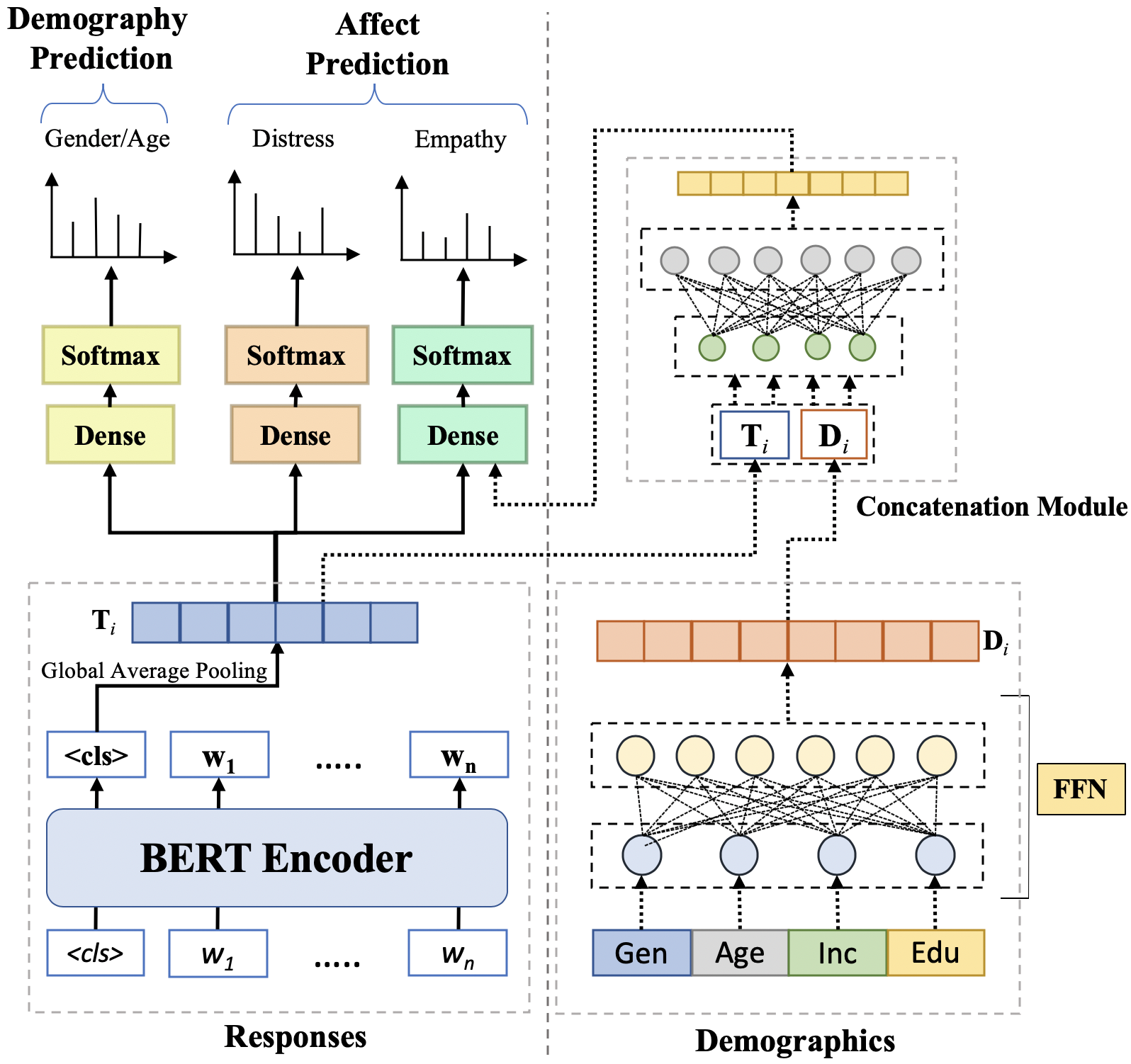}
    \caption{{\sc EmpathBERT} architecture.}
    \label{fig:overall_archi}
\end{figure}
Empathy assesses feelings of sympathy towards \emph{others}, and Distress measures anxiety and discomfort oriented towards \emph{self} \cite{davis1980interpersonal}.
Empathy has been positively associated to a number of well-being activities, such as volunteering \cite{batson1987distress}, charity \cite{pavey2012help}, and longevity \cite{poulin2013giving}, and in consumer marketing, advertising and customer interfaces \cite{wang2016experiencing, escalas2003sympathy}.
Works on empathy in text have focused on spoken dialogue, addressing conversational agents, psychological interventions, or call center transcripts \cite{Mcquiggan2007,Fung2016,Perez-rosas-etal-2017,Alam2018,Demasi2019}\nocite{Lin-etal-2019,Rashkin2019}.
\citet{Buechel-etal-2018} collected an empathy-distress dataset by leveraging users' reactions to textual stimulus content.
\citet{Sedoc2019} constructed an empathy lexicon by obtaining word ratings from document-level ratings from this dataset.
\citet{Xiao2012,Gibson2015,Khanpour-etal-2017} presented predictive models for empathy in the healthcare domain.
However, we believe none of the above works focus on (a) predicting empathy from textual reactions, and (b) studying the impact of demographics on the expression of empathy. 

Language preferences vary with user demographics \cite{Tresselt1964,Eckert2013,Garimella2016,Lin2018,Loveys2018}, and this has led to studies leveraging the user demographic information to obtain better language representations and classification models for various NLP tasks \cite{volkova-etal-2013-exploring,bamman-etal-2014-distributed,hovy-2015-demographic,Garimella-etal-2017}.
Owing to the recent success of large language models such as BERT \cite{devlin2018bert} and GPT-2 \cite{Radford2019} in improving the performances of several downstream tasks, we propose a BERT-based demographic-aware framework for empathy (distress) prediction, and through several comparative experiments, show that it surpasses existing baselines and demographic-agnostic approaches.

This paper makes three main contributions.
{\bf (1)} We present {\sc EmpathBERT}, a demographic-aware empathy (distress) prediction framework, using BERT-based models infused with demographic information.
{\bf (2)} Through comparisons against several baseline and demographic-agnostic approaches, we illustrate the importance of user demographics in end-to-end modeling and predicting empathy (distress).
{\bf (3)} Conversely, we show that empathy (distress) also contributes to demographic attribute prediction, by developing affect-aware models for demographic attribute prediction, backed by empirical comparison with baselines and generic models.
To the best of our knowledge, ours is the first computational effort addressing empathy (distress) through the lens of demographic biases, a phenomenon well-understood in psychology.   
\section{Dataset}
\label{sec:dataset}
We use the empathy-distress dataset introduced by \citet{Buechel-etal-2018}.
It consists of 418 news articles from popular news platforms, and responses to them from 403 annotators (5 articles each), resulting in a total of 2,015 responses.
Filtering the responses that deviated from the task description led to 1,860 responses (empathy: 916, distress: 905) , with a total token count of 173,686 (min: 52, max: 198, median: 84).
The number of responses per article ranges from 1 to 7, with an average of 4.46 responses per article. We report some example responses from the dataset in Table \ref{tab:empathy_scores}\footnote{Please refer to \citet{Buechel-etal-2018} for further details on the dataset.}.
We focus on the responses only, and use the empathy (distress) tags associated with these responses.
We group the data into binary classes for age ($\mathrm{C_0}$: $<35$, $\mathrm{C_1}$: $\geq$ 35), income ($\mathrm{C_0}$: $\leq\$50,000$, $\mathrm{C_1}$: $>\$50,000$), and education ($\mathrm{C_0}$: no degree, $\mathrm{C_1}$: bachelor's or above), to mitigate class imbalances.\footnote{We do not study race; it has even heavier class-imbalance.}
The resulting dataset is balanced for all dimensions, with a maximum deviation of $5.5\%$ (age) among classes. 

\begin{table*}[!t]
\centering
\scalebox{0.90}{
\begin{tabular}{p{12.4cm}p{3cm}p{1cm}}
\toprule
\textbf{\textsc{Text}} & \textbf{\textsc{Demographic Attributes}} & \textbf{\textsc{Score}}\\
\midrule
A 6.4 magnitude on the Richter scale earthquake has shaken up the whole capital of Santiago, Chile. Chile is very propense to earthquakes and natural disasters. We have heard of an earthquake that scaled out to be 8.8 and destroys over 200 thousand homes in Chile. I feel very bad for the people who died. and send out my compassion to the family of the 55 dead in this earthquake. & Female, Age $\geq$ 35, Education $<$ Bachelors, Income $\leq$ \$50,000 & 0.84\\
\midrule
This is just crazy, you have to feel for the mother, but at the same time what kind of apartment has that many violations and is still not punished.  They need to sue them and anybody involved with this.  I can’t believe that in today's society that tragedies like this are tolerated.  Somebody needs to go to jail for the death of this little girl and the injuries that her mother suffered. I can’t imagine what the mother is going through and she probably blames herself.  Things like this should just not happen. & Male, Age $\geq$ 35, Education $\geq$ Bachelors, Income $\leq$ \$50,000 & 0.82\\ 
\bottomrule
\end{tabular}
}
\caption{Qualitative examples of high empathy (above) and high distress (below) with scores on empathy and distress dimensions as predicted by our {\bf tBERT-C (fnn)} model.}
\label{tab:empathy_scores}
\end{table*}
\section{\textbf{\textsc{EmpathBERT}}}
\label{sec:framework}
In this section, we describe our approach for demographic-aware empathy (distress) prediction from text.
Figure \ref{fig:overall_archi} shows the proposed architecture.
Our model takes as input a response (a sequence of words $w_1, w_2, \dots, w_n$) and demographic information of the corresponding annotator.
We represent the response using BERT, a bidirectional Transformer-based \cite{vaswani2017attention} language model.
We use the final 768-dimensional hidden vector corresponding to the {\sc [CLS]} token as the aggregate sequence representation.
We employ cross-domain pre-training~\cite{sun2019fine}, fine-tuning, and multi-task fine-tuning~\cite{liu2019multi} techniques to customize BERT for our tasks.
\paragraph{Cross-domain Pre-training (PT).}
We use the pre-trained BERT language model trained on the English Wikipedia and Book Corpus \cite{Zhu2015} datasets for masked word and next sentence prediction, and perform further pre-training on demographic-specific datasets to introduce demographic-specific language preferences.
This enables slanting the BERT model towards a specific demographic group.
For this, we use a corpus different from the empathy dataset in two scenarios.
(1) \textsc{All}: train the BERT model on all of the external corpus, and (2) {\sc demographic-specific}: train only on the demographic-specific samples from the external corpus.

\paragraph{Fine-tuning Only (\textbf{tBERT}).}
BERT-based fine-tuning has had significant success, due to the ease in implementation and performance gains reported for various NLP tasks \cite{Huang2019,Liu2019}.
We fine-tune BERT for sequence classification by adding a classification layer, where the input is response represented by the hidden vector of the {\sc [CLS]} token, and output is the prediction for empathy (distress).
We train on generic data and demographic-specific portions, and compare the performances to study the demographic effect on empathy (distress) prediction. 
\paragraph{Multi-task Fine-tuning (tBERT-MT).} 
We fine-tune BERT in multi-task learning (MTL) setup for classification, similar to \cite{liu2019multi}, where the tasks under consideration are empathy (distress) classification and demographic attribute prediction.
Both the tasks have shared BERT layers, while the classification heads containing the final dense and softmax layers are specific to each task.
We replace the final dense and softmax layers in tBERT setup with multiple classification heads based on the number of tasks.
We experiment with
(1) {\it Alternative training}:
In each epoch, we cyclically train only one classification head, freezing the parameters of the remaining heads; and
(2) {\it Parallel training}: In each epoch, we train the model end-to-end on the joint loss from all the classification heads.\\
\B{Explicit Demographic Knowledge.}
PT, tBERT and tBERT-MT intrinsically infuse demographic information.
We also incorporate this explicitly by concatenating a {\it demographic vector $\vv{d}$} to the output of the global average pooling layer~\cite{lin2013network} from tBERT or tBERT-MT (concatenation in Figure \ref{fig:overall_archi}) in two ways.
(1) {\bf tBERT-[MT]-C}: $\vv{d}$ is a $d$-dimensional one-hot encoding vector ($d$: number of demographics).
(2) {\bf tBERT-[MT]-C (fnn)}: $\vv{d}$ is the output of a feedforward neural network (FNN), the input for which is a one-hot encoding vector.
Three dense layers
are stacked before the task-specific heads, and this model is trained end-to-end for empathy (distress) prediction.
In tBERT-MT where one of the tasks heads predicts a demographic attribute, the corresponding binary value in $\vv{d}$ is removed.
To assess the contribution of specific attributes, 
we also propose to concatenate a 1-bit encoding (\textbf{tBERT-[MT]-C (attribute)}) for each given attribute.  

\section{Experiments}
\label{sec:expt}
\begin{table*}
\centering
\scalebox{0.80}{
\begin{tabular}{|c|l|c|c|c|c|c|c|c|c|c|}
\toprule
&\textbf{Method$\rightarrow$}&\multicolumn{3}{c|}{PT}&\multicolumn{3}{c|}{tBERT}&\multicolumn{3}{c|}{PT + tBERT} \\ \hline 
&Test$\rightarrow$& M &F& $\mathrm{A_s}$  &M &F& $\mathrm{A_s}$  &M &F& $\mathrm{A_s}$ \\ \midrule 
\multirow{3}{*}{\rotatebox{90}{Empathy}}
&Male & {\bf 50.02$^\dagger$} & 52.42 & 62.70 & {\bf 64.73$^\ddagger$} & 60.37 & 62.22 & {\bf 61.82$^\dagger$} & 57.95 & 58.65 \\
&Female &  49.07& {\bf 53.12$^\star$} & 48.28 & 63.70 & {\bf 64.56$^\ddagger$} & 63.32 &  58.16 & {\bf 61.77$^\ddagger$} & 58.51 \\
&$\mathrm{All_s}$ & 49.74 & 52.91 & 49.64 & 63.08 & 62.19 & 63.00 & 57.24 & 58.63 & 56.30 \\
\midrule
\multirow{3}{*}{\rotatebox{90}{Distress}}
&Male  & {\bf 51.21$^\star$} & 52.26 & 52.41 & {\bf 64.44$^\ddagger$} & 61.56 & 62.11 & {\bf 61.92$^\dagger$} & 57.60 & 59.63 \\
&Female  & 50.71 & {\bf 52.77$^\star$} & 51.57 & 61.52 & {\bf 63.16$^\ddagger$} & 60.51  & 57.35& {\bf 59.30$^\dagger$} & 60.19 \\
&$\mathrm{All_s}$  & 49.43& 51.42& 50.53 & 63.18 & 62.77& 62.88  & 59.78 & 58.77 & 59.57 \\
\bottomrule
\end{tabular}}
\caption{{Accuracies using gender-specific training for empathy (distress) prediction. \B{M}ale, \B{F}emale, \B{$A_s$}ll denote the respective data subsets. $A_s$ is a sampled dataset with approximately equal number of samples from $M$ and $F$ subsets, hence comparable in size.}\footnotemark} 
\label{tab:genderSpecificTraining}
\end{table*}

\begin{table*}
\centering
\scalebox{0.80}{
\begin{tabular}{|c|l|c|c|c|c|c|c|c|c|c|}
\toprule
&\textbf{Dem$\rightarrow$}&\multicolumn{3}{c|}{Age}&\multicolumn{3}{c|}{Income}&\multicolumn{3}{c|}{Education} \\ \midrule 
&Test$\rightarrow$ & $\mathrm{C_0}$ & $\mathrm{C_1}$ & $\mathrm{A_s}$  & $\mathrm{C_0}$ & $\mathrm{C_1}$ & $\mathrm{A_s}$ & $\mathrm{C_0}$ & $\mathrm{C_1}$ & $\mathrm{A_s}$ \\ \midrule 
\multirow{3}{*}{\rotatebox{90}{Empathy}}
& $\mathrm{Class_0}$  & {\bf 62.79$^\star$} & 62.59 & 61.44  & {\bf 62.05$^\ddagger$} & 61.82 & 62.66  & 59.44 & 61.18 & 58.81 \\
& $\mathrm{Class_1}$  & 59.27 & {\bf 64.95$^\star$} & 60.05 & 58.40 & {\bf 64.96$^\ddagger$} & 60.41 & {\bf 62.34$^\ddagger$} & {\bf 63.40$^\ddagger$} & 61.40 \\
& $\mathrm{All_s}$ & 59.73 & 62.05 & 60.26 & 60.62 & 63.21 & 60.92  & 60.81& 63.03 & 59.38 \\
\midrule
\multirow{3}{*}{\rotatebox{90}{Distress}}
& $\mathrm{Class_0}$  & {\bf 62.80$^\ddagger$} & 62.32 & 61.01  & {\bf 62.46$^\ddagger$} & 61.43 & 60.86 & {\bf 62.65$^\ddagger$} & 62.04 & 62.30 \\
& $\mathrm{Class_1}$  & 57.20 & {\bf 68.08$^\ddagger$} & 60.68  & 60.23 & {\bf 66.59$^\dagger$} & 62.39 &  60.45 & {\bf 66.85$^\dagger$} & 63.31\\
&$\mathrm{All_s}$  & 60.89 & 65.08& 61.16 & 59.92 & 61.54 & 60.88 &61.80 & 63.13 & 62.06 \\ 
\midrule
\end{tabular}}
\caption{{Demographic-specific training accuracies for empathy (distress) prediction.}}
\label{tab:demoSpecificTraining}
\end{table*}
\begin{table*}[t]
\centering
\scalebox{0.75}{
\begin{tabular}{|c|l|cc|cc|cc|cc|cc|cc|}
\toprule
&\textbf{Affect$\longrightarrow$}&\multicolumn{6}{c|}{Empathy}&\multicolumn{6}{c|}{Distress}\\ \midrule
&Test Set$\longrightarrow$& \multicolumn{2}{c|}{All}&\multicolumn{2}{c|}{Male}& \multicolumn{2}{c|}{Female} & \multicolumn{2}{c|}{All}& \multicolumn{2}{c|}{Male} & \multicolumn{2}{c|}{Female} \\ \midrule
\multirow{5}{*}{\rotatebox{90}{Trad.ML}} &\textbf{Approach$\downarrow$} & \textbf{$F_1$} & \textbf{Ac}& \textbf{$F_1$} & \textbf{Ac}& \textbf{$F_1$} & \textbf{Ac} & \textbf{$F_1$} & \textbf{Ac}& \textbf{$F_1$} & \textbf{Ac}& \textbf{$F_1$} & \textbf{Ac} \\ 
\cmidrule{2-14}

&RF-text & 57.5 & 59.9 & 58.4 & 60.3 & 56.4 & 57.5 & 58.4 & 61.0 & 58.9 & 60.9 & 57.9 & 61.3\\ 
&RF-dem & 58.99 & 59.12 & 58.59 & 58.64 & 59.77 & 59.77 & 58.06 & 58.03 & 60.61 & 60.70 & 59.77 & 59.29\\ 
&RF-text$+$dem & 58.5 & 60.7 & 57.9 & 59.7 & 57.1 & 59.7 & 58.4 & 60.5 & 59.6 & 59.9 & 58.1 & 61.2\\ \midrule

\multirow{4}{*}{\rotatebox{90}{DL Models}} &CNN&  59.5 & 61.3 & 60.7 & 62.1 & 58.2 & 60.5 & 58.8 & 63.9 & 57.8 & 62.5 & 59.9 & 63.5\\
&biLSTM & 53.3 & 55.4 & 55.4 & 57.1 & 50.8 & 53.5 & 57.1 & 59.3 & 54.3 & 56.9 & 60.2 & 62.1  \\
 &biLSTM-Attention & 60.8 & 62.6 & 60.0 & 62.0 & 61.7 & 63.3 & 59.9 & 62.7 & 59.8 & 62.3 & 59.8 & 63.1 \\ 
 &BERT & 65.6 & 49.0 & 65.3 & 48.8 & 65.9 & 49.2 & 66.1 & 49.5 & 65.8 & 49.3 & 66.3 & 49.6\\
\cmidrule{1-14}
\multirow{13}{*}{\rotatebox{90}{Proposed Methods}}&Aff-biLSTM-text$+$dem & 61.9 & 63.0 & 63.0 & 63.6 & 60.8 & 62.3 & 62.9 & 64.2 & 62.9 & 64.6 & 62.9 & 63.7 \\
&tBERT (E)& \textbf{67.1$^\dagger$} & \textbf{67.8$^\dagger$} & \textbf{68.7$^\star$} & \textbf{69.4$^\star$} & \textbf{65.4$^\star$} & \textbf{66.1$^\star$} & -- & -- & -- & -- & -- & -- \\
&tBERT (D)& -- & -- & -- & -- & -- & -- & 67.6 & 67.0 & 69.3 & 68.6 & 65.7 & 65.1 \\
\cmidrule{2-14}
&tBERT-MT-(E$+$D)  & 65.2 & 66.2 & 66.7 & 67.6 & 63.6 & 64.6 & \textbf{69.2$^\ddagger$} & \textbf{68.5$^\ddagger$} & 71.2 & 70.4 & \textbf{66.7$^\dagger$} & \textbf{ 66.3$^\dagger$}  \\
&tBERT-MT-G &(E) 63.9 & 64.9 & 65.0 & 66.7 & 62.8 & 62.8 &(D) 67.0 & 67.5 & 70.8 & 70.3 & 62.3 & 64.3\\
&tBERT-MT-(E$+$D)-G & 64.5 & 64.7 & 65.7 & 66.3 & 63.1 & 63.0 & 68.1 & 67.7 & \textbf{71.3$^\star$} & \textbf{70.5$^\star$} & 64.3 & 64.5 \\
\cline{2-14}
&tBERT-MT-A  & (E) 61.8 & 63.5 & 65.3 & 66.7 & 58.1 & 60.0 & (D) 65.0 & 65.1 & 67.6 & 67.5 & 62.2 & 62.5 \\
&tBERT-MT-(E$+$D)-A & 64.1 & 65.2 & 65.8 & 66.8 & 62.3 & 63.5 & 66.0 & 66.2 & 69.3 & 68.8 & 62.1 & 63.1 \\
\cmidrule{2-14}
&tBERT-C (fnn)   &(E) 66.4 & 67.4 & 67.6 & 68.6 & 65.0 & 66.0 & 67.4 & 67.4 & 69.4 & 69.2 & 65.0 & 65.3 \\
&tBERT-C   &  66.0 & 66.4 & 66.8 & 67.0 & 65.0 & 65.8 & 68.2 & 67.7 & 69.9 & 69.5 & 66.2 & 65.6 \\
&tBERT-C (gender)   &  64.3 & 66.8 & 65.0 & 67.7 & 63.5 & 65.9 & 66.8 & 66.9 & 68.6 & 68.7 & 64.7 & 64.9 \\
&tBERT-MT-G-C& (E) 63.8 & 66.0 & 65.4 & 67.6 & 62.2 & 64.2 &(D) 65.9 & 67.0 & 68.6 & 68.9 & 62.5 & 64.9\\
&tBERT-MT-(E$+$D)-G-C& (E) 62.2 & 64.0 & 64.5 & 65.7 & 59.7 & 62.2 &(D) 64.6 & 66.1 & 67.5 & 68.3 & 61.3 & 63.6 \\
\bottomrule
\end{tabular}}
\caption{\label{font-table} {Demographic-aware empathy (distress) prediction. For tBERT-MT, the multitask attributes are specified in the method name i.e. gender (-G), age (-A) along with empathy (E) or distress (D) along side the accuracies. F$_{1}$: F1 score; Ac: Accuracy.}}
\label{tab:empathy_prediction_results}
\end{table*}

\begin{table}[]
    \centering
    \scalebox{0.58}{
    \begin{tabular}{|l|c|c|c||c|c|c|}
    \toprule
    Demography $\longrightarrow$ & \multicolumn{3}{c||}{Gender} & \multicolumn{3}{c|}{Age} \\ \cmidrule{2-7 }
    Dataset $\longrightarrow$& All & Em &Dist &All & Em & Dist \\ \midrule
    RF-text & 59.8 & 60.8 & 58.7  & 56.5  & 55.7 & 57.3\\
    RF-text-E/D &  58.0 &  59.1 &  56.9&  56.6 &  54.2  & 59.1\\ \midrule
    Aff-biLSTM(att)-text  & 59.2  & 60.1  & 58.2  & 56.2  & 57.7  & 54.6\\
    Aff-biLSTM(att)-text-E/D  & 58.9  & 60.2  & 57.4  & 56.9  & 57.3 & 56.6\\ 
    BERT  & 47.5 & 47.3 & 47.7  & 40.5 & 41.3  & 39.6 \\ \midrule
    tBERT& (G) 64.2$^\ddagger$  & 65.2  & 63.4 & (A) 62.7$^\star$ &  63.2$^\ddagger$  & 63.8$^\ddagger$ \\
    tBERT-MT-E &  62.0 & 61.5 & 63.3 & 60.1  & 61.1 & 61.9\\
    tBERT-MT-D & 61.6 & 61.7 & 63.9 & 60.6  & 60.8 & 61.7\\
    tBERT-MT-(E$+$D)  & 63.1 & 62.9 & 65.1$^\star$ & 61.6  & 59.8 & 63.7\\ \bottomrule
    \end{tabular}}
    \caption{F1 values of affect-aware demography prediction.}
    \label{tab:affectAwareDemPrediction}
\end{table}
\footnotetext{Statistical significance using McNemar's Test \cite{mcnemer} with $^\star$ $p<0.05$, $^\dagger$ $p<0.01$, $^\ddagger$ $p<0.001$.}
We model empathy (distress) prediction as a binary classification task.
To study the efficacy of empathy (distress) to predict demography attributes, we also conduct experiments for empathy (distress)-aware demographic attribute prediction. 
Such a prediction can be used for further demographic removal from text to mitigate adversarial attacks and protect privacy of users \cite{elazar-goldberg-2018-adversarial}.\\
\B{Implementation Details}
(1) Cross-domain Pre-training: We use the Blog Authorship Corpus\footnote{\url{https://u.cs.biu.ac.il/~koppel/BlogCorpus.htm}} \cite{schler2006effects}, which consists of 681,288 blogposts and self-provided demographic attributes, gender, age, industry, and astrological sign of the corresponding 19,320 bloggers to further pre-train BERT. Out of these we use the gender attribute to pre-train for male-specific and female-specific pre-training experiments. We train the model on the \textit{Masked Language Model task}~\cite{taylor1953cloze} for 10 epochs using a learning rate of 3e-5. (2) Finetuning: We train the model end-to-end (110M parameters) using binary cross-entropy loss and decoupled weight decay Adam optimizer~\cite{loshchilov2017decoupled}, in batches of 32. The best performance is observed when the maximum input sequence length is set to 150, learning rate to 3e-5, and number of epochs to 3.
(3) Explicit Demographic Attributes: We use gender, age, education and income attributes corresponding to each annotator in the empathy dataset.
The $d$-dimensional vector  size $4$ resulting in a $16$-d FFN output. \\
\B{Evaluation metrics.}
We use five-fold cross validation
(five random shuffled restarts) with 80-20 train-test proportions,
and report the F1 and accuracy (Ac) averaged across the 5 runs on the test set.\\
\B{Baselines.}
We compare our model against the Random Forest (RF) model with Glove embeddings \cite{Pennington-etal-2014} for text and demographic attributes (excluding the prediction attribute) as one-hot vectors as features. We also report performance against deep learning baselines, CNN \cite{kim-2014-convolutional}, biLSTM, and biLSTM with Attention \cite{Yang2016HierarchicalAN} and the pre-trained BERT without further training.
\subsection{Results}
Table \ref{tab:genderSpecificTraining} shows the accuracies using BERT for pretraining (PT), fine-tuning (tBERT), and both (PT + tBERT) for gender-specific empathy (distress) prediction.
On the $M$ and $F$ test sets, models trained on the same demographic subset ($M$ or $F$) outperform those trained on the opposite subset or $A_s$.
The acccuracies of plain BERT are
$48.37$, $49.49$, and $50.42$
on $A_s$, $M$, and $F$ test sets respectively for empathy prediction. tBERT outperforms all other variants.
The results support our hypothesis that empathy is dependent on and influenced by the gender associated with the author. 
We note similar patterns for age, income, and education
(Table \ref{tab:demoSpecificTraining}).

Table \ref{tab:empathy_prediction_results} shows results for empathy (distress) prediction using tBERT-[MT]-[C (fnn/attribute)] variants trained on the full dataset. 
In the notation, we replace [MT] with the heads on which the multi-tasking is performed. For example tBERT-MT-(E+D)-G-C implies fine-tuned BERT with empathy prediction, distress prediction, and gender prediction multi-tasking heads with demographic information concatenated to the text representation directly before classification.\footnote{In the models where a demographic attribute prediction is involved, we remove that attribute from the demographic vector.} 
We report performances on demographic-wise test sets ($A$, $M$, $F$). 
\\
\B{Insights:} (1) \B{tBERT} variants with a single training objective outperform all baselines.
(2) Performance of \B{tBERT-MT} varies with the affect dimension.
Empathy prediction shows marginal loss in performance with explicit concatenation (\B{tBERT-C}) and further loss in the multitask setup.
(3) For distress, introduction of gender as the demographic attribute shows an observable improvement across different test sets. (4) A similar trend is observed for age.
Table \ref{tab:affectAwareDemPrediction} shows performance of age and gender prediction with empathy (distress)-aware models on affect-wise test sets (Empathy (Em) and Distress (Dist)).
Empathy-aware gender prediction models show consistent improvement over baselines, with \B{tBERT (G)} reporting the best score when tested on the complete dataset and empathy-specific test set. 
\B{tBERT (A)} helps improve the accuracies for age prediction by atleast 5\% over baselines for the complete (All) test set.
For the empathy-specific test set, best results are observed with  MTL (\B{tBERT-MT-(E+D)}). We infer that while having affect-aware demographic prediction models do improve performance over fine-tuned models, they may also lead to a marginally negative impact. 
The overall inference from above experiments is that demographic-aware models aid affect predictions but the reverse relationship is much weaker.
End-to-end training across a variety of train sets and demographic attributes establishes that the variance observed in language preferences and expressions has an impact on the manner of expressing empathy and distress in reactions. 
\section{Conclusion}
\label{sec:conclusion}
We proposed a novel demographic-aware empathy prediction framework based on fine-tuning and multi-tasking using BERT, showed that it surpasses existing methods, and illustrated the impact of demography in modeling subjective phenomena such as empathy and distress.
Our framework is generalizable, and we extended it to empathy-aware demography prediction, and showed that empathy also improves demographic prediction.
We believe this is a significant checkpoint towards developing models for empathy (distress), and tapping into demographic information while doing so.
\bibliographystyle{acl_natbib}
\bibliography{eacl2021}
\end{document}